\documentclass{article}

     \usepackage[preprint]{neurips_2021}

\usepackage[utf8]{inputenc} 
\usepackage[T1]{fontenc}    
\usepackage{hyperref}       
\usepackage{url}            
\usepackage{booktabs}       
\usepackage{amsfonts}       
\usepackage{nicefrac}       
\usepackage{microtype}      
\usepackage{xcolor}         
\usepackage{mathptmx} 
\usepackage{amsmath}
\usepackage{bm}
\usepackage{amsthm}
\usepackage{centernot}
\usepackage{amssymb}
\usepackage{dsfont}

\newtheorem{theorem}{Theorem}[subsection]

\newtheorem{proposition}[theorem]{Proposition}
\newtheorem{corollary}[theorem]{Corollary}

\DeclareMathAlphabet{\mathcal}{OMS}{cmsy}{m}{n}

\DeclareMathOperator*{\argmin}{argmin}   

\title{Model Mis-specification and Algorithmic Bias}

\author{%
    Runshan Fu, Yangfan Liang, Peter Zhang \\
  Heinz College of Information Systems and Public Policy\\
  Carnegie Mellon University\\
  Pittsburgh, PA 15213 \\
  \texttt{runshanf, yangfanl, yunz2@andrew.cmu.edu}
}

\begin{document}

\maketitle

\begin{abstract}
Machine learning algorithms are increasingly used to inform critical decisions. There is a growing concern about bias, that algorithms may produce uneven outcomes for individuals in different demographic groups. In this work, we measure bias as the difference between mean prediction errors across groups. We show that even with unbiased input data, when a model is mis-specified: (1) population-level mean prediction error can still be negligible, but group-level mean prediction errors can be large; (2) such errors are not equal across groups; and (3) the difference between errors, \emph{i.e.}, bias, can take the worst-case realization. That is, when there are two groups of the same size, mean prediction errors for these two groups have the same magnitude but opposite signs. In closed form, we show such errors and bias are functions of the first and second moments of the joint distribution of features (for linear and probit regressions). We also conduct numerical experiments to show similar results in more general settings. Our work provides a first step for decoupling the impact of different causes of bias.
\end{abstract}

\section{Introduction} \label{sec.intro}

\subsection{Motivation and research questions}

Consider a bank that uses a machine learning model to make loan approval and interest rate decisions. If individuals from one demographic group are systematically assigned higher risk scores than they deserve, and individuals from another group are systematically assigned lower risk scores than they deserve, then this may be undesirable. Our goal in this work is to understand if and how such discrepancy is due to the algorithmic part of the machine learning process.
More specifically, we want to answer these questions: How do we define such discrepancy? When would a machine learning model create discrepant outcomes for different groups of individuals? And when that happens, can we quantify the discrepancy? 

There is a plethora of perspectives to study these questions. To refine the scope of our analysis, we make two assumptions. First, we assume that the data used to train the machine learning model represent the true underlying distribution of feature and outcome variables, \emph{e.g.}, there is no sampling bias or labeling error. Second, we assume that the machine learning model produces a risk score for each individual -- and in the example above, it is in turn used to make (binary) loan approval decisions and (continuous or discrete) interest rate decisions -- but we focus on risk score only, not the downstream decisions such as threshold-setting in common classification tasks.

By assuming unbiased data and focusing on the direct output of a machine learning model, we remove the effects of upstream and downstream human influence.
That is \emph{not} to say that data biases and human influence are not important in the study of machine learning bias and (un)fairness. On the contrary, they are crucial, and that is precisely why we separate them. Our ``open-box'' investigation could help attribute unfairness to data, algorithms, and humans more precisely in future research. 

\subsection{Definitions}

We denote the outcome variable that the machine learning model aims to predict as $Y$, and assume that in the true data generating process (DGP), $Y$ is a function of the \textbf{}feature vector $\mathbf{X} = (X_1, X_2, ..., X_n)^\intercal$:
\begin{align}
    \label{eq.true.data.generating.process}
    Y = h(\mathbf{X}).
\end{align}

We use $\phi$ to denote a trained machine learning model, and the \emph{prediction error} is defined as
\begin{align*}
    e = \hat{Y}  - Y  = \phi(\mathbf{{X}}\textbf{}) - Y.
\end{align*}
In this paper, we partition our results into two cases: $\phi$ is \emph{correctly specified}, and $\phi$ is \textit{mis-specified} (when compared with $h$). We further differentiate mis-specification into \emph{mismatch between function classes} and \emph{variable omission}. For example, if $h$ is a second-degree polynomial function, and $\phi$ is drawn from the family of linear functions, then $\phi$ is mis-specified in the sense of function class mismatch. If $\mathbf{X}_{\text{short}}$ is a feature vector that contains a subset of the original features, and $\phi$ only depends on $\mathbf{X}_{\text{short}}$, then $\phi$ is mis-specified in the sense of variable omission. For the latter, we also slightly abuse the notation and write $\phi(\mathbf{X}_{\text{short}})$ instead of $\phi(\mathbf{{X}})$.


We denote the \emph{population-level mean prediction error} as $b$,
\begin{align*}
    b(\phi) = \mathbb{E}(e|\phi).
\end{align*}
Prediction error $e$ is closely related to the concept of ``bias'' in bias-variance trade-off, as $\hat{Y}$ can be viewed as the expected prediction of $Y$ over different possible training sets given by a machine learning model $\phi$. 

For exposition, we assume there are two groups of individuals: a protected group and a regular group, but many results can directly generalize to an arbitrary number of groups. 
The protected group is the group of individuals that have been historically discriminated against and need to be protected (\emph{e.g.}, females or African Americans), and the other individuals belong to the regular group. We use $A$ to denote the group attribute (or sensitive attribute), where $A=1$ represents the protected group and $A=0$ represents the regular group. With a slight abuse of notation, the \textit{group-level mean prediction errors} are defined as $b(\phi, A=0) = \mathbb{E}(e|\phi, A=0)$ and $b(\phi, A=1)=\mathbb{E}(e|\phi, A=1)$, respectively.

We define \textit{outcome bias attributable to algorithm} (more simply, \emph{algorithmic bias} or \emph{outcome bias}) as
\begin{align*}
    \tau(\phi)= b(\phi, A=1) - b(\phi, A=0).
\end{align*}
In other words, bias is defined as the systematic difference in prediction errors between the two groups.
For example, if $Y$ is individual creditworthiness, and for a machine learning model, $b(\phi, A=1) > b(\phi, A=0)$, then this suggests that the machine learning model produces systematically higher creditworthiness scores for the individuals in the regular group, compared to those in the protected group. When $Y$ is a desirable variable (\emph{i.e.}, higher $Y$ leads to more favorable decisions), then a positive $\tau$ suggests bias against the protected group, and a negative $\tau$ suggests bias against the regular group. If $\tau(\phi) = 0$, then $\phi$ is considered unbiased (or fair). As will be shown in Sections \ref{sec.regression} and \ref{sec.classification}, the direction and magnitude of bias nontrivially depends on the joint distribution of feature variables.


\subsection{Machine learning fairness literature}

Previous works on fair machine learning have studied various problems related to the definition, detection, mitigation, and sources of bias. Fairness notions in the current literature can be broadly categorized into four groups: unawareness \citep{dwork2012fairness}, individual fairness \citep{dwork2012fairness}, group fairness \citep{10.5555/3157382.3157469, mehrabi2019survey, 10.1145/3376898}, and counterfactual fairness \citep{NIPS2017_a486cd07}. It has been shown that many fairness notions are incompatible with each other, and the decision makers should carefully choose fairness notion based on the specific scenarios \citep{pedreshi2008discrimination, chouldechova2017fair, kleinberg2016inherent, berk2018fairness}. In this paper, we provide a new measure of algorithmic bias that separates bias directly caused by machine learning model from data biases and human influence. For a detailed discussion and a proof of how our definition of bias relates to the existing notions of machine learning (un)fairness, we refer the readers to the Appendix. A general view in the literature is that algorithms are not biased \emph{per se}, but they may reflect and amplify data biases. Such data biases include biased labels \citep{barocas2016big, obermeyer2019dissecting}, imbalanced group representation \citep{10.1145/3376898}, and data quality disparity \citep{barocas2016big}. In this paper, we demonstrate that model mis-specification can lead to outcome bias even if data is not biased. 


\subsection{Summary of results}

First, we prove that when a model is correctly specified, the mean prediction errors and bias are zero. This implies that if there is redundant encoding in the data, \emph{i.e.}, when features used in prediction are correlated with the sensitive attributes, the model needs not produce biased outcome.

We then quantify errors and biases when a model is mis-specified with omitted variables. In particular, we show that group-level mean prediction errors can be large even if population-level mean prediction error is small. Such errors are different across groups, leading to bias. The magnitude of such bias increases as the model mis-specification worsens. In general, mis-specification due to omitted variables and mis-specification due to the mismatch of function classes both lead to this outcome. We quantify such errors and bias in closed-form for linear and probit regressions, and conduct numerical experiments to show them empirically in a broad set of settings.

We divide our analysis into three parts. Section \ref{sec.regression} proves the aforementioned results for linear regression under variable omission. Section \ref{sec.classification} proves the results for binary classification under variable omission. Section \ref{sec.numerical} presents numerical results for a broader set of data generating processes and models under various model mis-specifications.

\section{Regression} \label{sec.regression}

In this section, we demonstrate the absence (presence) of bias when the model is correctly (mis-) specified in linear regressions. Specifically, we first show that when the regression model is correctly specified, the population-level and group-level mean prediction errors and bias are zero. Next, we show that if some features are omitted, a common type of mis-specification, then the model produces different group-level mean prediction errors, thus the bias is nonzero. The main analytical task here is to quantify such errors and biases in closed-form. As we will see, these quantities are closely associated with the joint distribution of feature variables. In other words, we quantify bias ($\tau$) using first and second moment information of relevant features. Throughout this section, we study linear regression with two feature variables, one of which is subject to omission. Analogous results can be developed for general multiple regression.

\subsection{Data generating process}
For simplicity, we assume that there are two relevant random variables, $X_1$ and $X_2$, and $h$ is a linear function:
\begin{align*}
h(\mathbf{X}) = \beta_0 + \beta_1 X_1 + \beta_2 X_2 + \epsilon,
\end{align*}
where $\epsilon$ is random noise with $\mathop{\mathbb{E}} (\epsilon | \mathbf{X}) = 0$ and $\mathop{\mathbb{E}}(\epsilon | A=1) = \mathop{\mathbb{E}}(\epsilon | A=0) = 0$. Let $\bm{\beta} = (\beta_0, ~\beta_1, ~\beta_2)$, and $\mathbf{X} = (1, ~X_1, ~X_2)$ with a slight abuse of notation, then the true data generating process can be written as
\begin{align*}
Y = \mathbf{X}^\intercal \bm{\beta} + \epsilon.
\end{align*}

\subsection{Correctly specified model}

The next two results show that when the model is correctly specified, population-level mean prediction error is zero (Proposition \ref{prop:regression_pop_zero}), and algorithmic bias is zero (Corollary \ref{cor:regression_tau_zero}). In particular, this is true even if the protected group is systematically different from the regular group in the data, \emph{e.g.}, protected group having more negative labels; or protected group being much smaller in size than the regular group.

\begin{proposition}\label{prop:regression_pop_zero}
    When the model is correctly specified, the mean prediction error in the population equals zero. That is, if $\xi$ is a random error with $\mathop{\mathbb{E}} (\xi | \mathbf{X}) = 0$ and $ \phi(X) = \mathbf{X}^\intercal \bm{\gamma} + \xi$, then $\mathop{\mathbb{E}}(e) = 0.$
\end{proposition}

\begin{proof}

$\bm{\gamma}$ is estimated through the following criteria
\begin{equation}
\begin{aligned}
    \label{eq.best.linear.predictor}
    \bm{\gamma} = \argmin_{\bm{\gamma}'} \mathop{\mathbb{E}} \left[ (Y-\mathbf{X}^\intercal \bm{\gamma}')^2 \right].
\end{aligned} 
\end{equation}

We can easily find that the solution to Eq.\ref{eq.best.linear.predictor} satisfies the following condition:
\begin{equation}
\begin{aligned}
    \label{eq.sol.best.linear.predictor}
    \bm{\gamma} &= ( \mathop{\mathbb{E}} (\mathbf{X} \mathbf{X}^\intercal) )^{-1} \mathop{\mathbb{E}} (\mathbf{X}Y ) =  ( \mathop{\mathbb{E}} (\mathbf{X} \mathbf{X}^\intercal) )^{-1} \mathop{\mathbb{E}} (\mathbf{X} ( \mathbf{X}^\intercal \bm{\beta} + \epsilon) ) \\
    &= \bm{\beta}.
\end{aligned}
\end{equation}

Thus, the best linear projection of $Y$ given $\mathbf{X}$ is
\begin{align}
    \label{eq.best.linear.projection}
    \hat{Y} = \mathbf{X}^\intercal \bm{\gamma} =  \mathbf{X}^\intercal \bm{\beta}.
\end{align}

By Eq.\ref{eq.best.linear.projection}, the prediction error $e =\hat{Y} - Y$ would be equal to the negative random noise $\epsilon$
\begin{align*}
    e = \hat{Y} - Y = \mathbf{X}^\intercal \bm{\beta} - Y = - \epsilon.
\end{align*}

Therefore, we have
\begin{align*}
    \mathop{\mathbb{E}}(e) = \mathop{\mathbb{E}}(-\epsilon) = 0.
\end{align*}
\end{proof}

\begin{corollary}\label{cor:regression_tau_zero}
When the model is correctly specified, the algorithmic bias equals zero. That is, $\tau = \mathop{\mathbb{E}}(e | A=1) - \mathop{\mathbb{E}}(e | A=0) = \mathop{\mathbb{E}}(\epsilon | A=1) - \mathop{\mathbb{E}}(\epsilon | A=0) = 0$.
\end{corollary}
This result follows directly from assumptions.

Consequently, a common belief that redundant encoding is problematic is incomplete. Here we can see that as long as the model is correctly specified, there needs not be any bias, even if feature values are correlated with the sensitive attribute $A$.

Some may argue that perfect model specification rarely exists, therefore the aforementioned results have little practical implication. But as we will show in the upcoming results, this line of inquiry allows us to provide some ``open-box'' analysis to understand machine learning unfairness. This ultimately will help attribute causes of bias more accurately to data, algorithms, and human influence.

\subsection{Model mis-specification and bias}

The following two results show that when the model is mis-specified with omitted variables, population-level mean prediction error equals zero (Proposition \ref{prop:regression_omitted_pop_zero}), but the difference between the mean prediction errors of the two groups would not equal zero (Proposition \ref{thm:regression_omitted_tau_not_zero}). In particular, when two groups have equal size, this would lead to the worst-case scenario for the difference between the mean prediction errors of the two groups (Corollary \ref{cor:regression_omitted_worst_case}).

Consider the case when the model is mis-specified with omitted variables ($X_2$ is omitted from the model). Here the estimation model is
\begin{align*}
    Y = \mathbf{X}_{\text{short}}^\intercal \bm{\gamma}_{\text{short}} + u,
\end{align*}
where $\mathbf{X}_{\text{short}} = (1, X_1)$, $\bm{\gamma}_{\text{short}}=(\gamma_0, \gamma_1)$, and $u$ is a random error with $\mathop{\mathbb{E}} (u | \mathbf{X}_{\text{short}}) = 0$. We denote the prediction error as $e_{\text{short}}$:
\begin{align*}
    e_{\text{short}}= \hat{Y}_{\text{short}} - Y.
\end{align*}

\begin{proposition}
    \label{prop:regression_omitted_pop_zero}
    When the model is mis-specified with omitted variables, the mean prediction error in the population equals zero, \emph{i.e.}, $ \mathop{\mathbb{E}}(e_{\text{short}}) = 0$.
\end{proposition}

\begin{proof}
Similar to Eq.\ref{eq.sol.best.linear.predictor}, here $\bm{\gamma}_{\text{short}}$ is derived as follows:
\begin{align*}
    \bm{\gamma}_{\text{short}} &= ( \mathop{\mathbb{E}} (\mathbf{X}_{\text{short}} \mathbf{X}_{\text{short}}^\intercal) )^{-1} \mathop{\mathbb{E}} (\mathbf{X}_{\text{short}} Y ) \\
    &= \frac{1}{\mathop{\mathbb{E}}(X_1^2) - (\mathop{\mathbb{E}} X_1)^2 } 
    \begin{pmatrix} \beta_0 (\mathop{\mathbb{E}}(X_1^2) - (\mathop{\mathbb{E}} X_1)^2 ) + \beta_2 ( \mathop{\mathbb{E}} X_1^2 \mathop{\mathbb{E}} X_2 - \mathop{\mathbb{E}} X_1 \mathop{\mathbb{E}} X_1 X_2) \\
    \beta_1 ( \mathop{\mathbb{E}}(X_1^2) - (\mathop{\mathbb{E}} X_1)^2) + \beta_2 ( \mathop{\mathbb{E}} X_1 X_2 - \mathop{\mathbb{E}} X_1 \mathop{\mathbb{E}} X_2 )
    \end{pmatrix}.
\end{align*}

Thus, we know that:
\begin{align*}
    \gamma_0 &= \beta_0 + \frac{\mathop{\mathbb{E}} X_1^2 \mathop{\mathbb{E}} X_2 - \mathop{\mathbb{E}} X_1 \mathop{\mathbb{E}} X_1 X_2}{Var(X_1)} \beta_2, \\
    \gamma_1 &= \beta_1 + \frac{Cov(X_1, X_2)}{Var(X_1)} \beta_2.
\end{align*}

The prediction error in this model would be as follows:
\begin{equation}
\begin{aligned}
    e_{\text{short}} &= \hat{Y}_{\text{short}} - Y =  (\gamma_0 + \gamma_1 X_1 ) - (\beta_0 + \beta_1 X_1 + \beta_2 X_2 + \epsilon) \\
    &= (\gamma_0 - \beta_0) + \frac{Cov(X_1, X_2)}{Var(X_1)} \beta_2 X_1 - \beta_2 X_2 - \epsilon.
    \label{eq.linear.omitted.residual}
\end{aligned}
\end{equation}

Therefore we have:
\begin{align*}
    \mathop{\mathbb{E}} (e_{\text{short}}) &=  \frac{\mathop{\mathbb{E}} X_1^2 \mathop{\mathbb{E}} X_2 - \mathop{\mathbb{E}} X_1 \mathop{\mathbb{E}} X_1 X_2}{Var(X_1)} \beta_2 + \frac{Cov(X_1, X_2)}{Var(X_1)} \beta_2 \mathop{\mathbb{E}} X_1 - \beta_2 \mathop{\mathbb{E}} X_2 = 0 .
\end{align*}
\end{proof}

\begin{proposition}
When the model is mis-specified with omitted variables, the group-level mean prediction errors would not equal zero, unless for each feature, its conditional expectation (on $A$) is the same for $A=0$ and $A=1$. That is $ \mathop{\mathbb{E}}(e_{\text{short}} | A=a) \neq 0 $ where $a=0$ or $1$, unless $\mathbb{E}(X_1 |A=a) = \mathbb{E} X_1$ and $\mathbb{E}(X_2 |A=a) = \mathbb{E} X_2$ $\forall a $.
\end{proposition}

\begin{proof}


From Eq. \ref{eq.linear.omitted.residual} we have the following:
\begin{align*}
    & \mathop{\mathbb{E}}(e_{\text{short}} | A=a) \\
    &=  \frac{\mathop{\mathbb{E}} X_1^2 \mathop{\mathbb{E}} X_2 - \mathop{\mathbb{E}} X_1 \mathop{\mathbb{E}} X_1 X_2}{Var(X_1)} \beta_2 + \frac{Cov(X_1, X_2)}{Var(X_1)} \beta_2 \mathop{\mathbb{E}} (X_1 | A=a) - \beta_2 \mathop{\mathbb{E}} (X_2 | A=a) \\
    & \neq 0, 
\end{align*}

unless $\mathbb{E}(X_1 |A=a) = \mathbb{E} X_1$ and $\mathbb{E}(X_2 |A=a) = \mathbb{E} X_2$, $\forall a = 0, 1$.

\end{proof}

\begin{proposition}

\label{thm:regression_omitted_tau_not_zero} 

When the model is mis-specified with omitted variables, the difference between the group-level mean prediction errors would not equal zero and the model is biased under our fairness notion, unless the group-specific mean of features satisfies a specific condition. That is, $ \tau = \mathop{\mathbb{E}}(e_{\text{short}} | A=1)  - \mathop{\mathbb{E}}(e_{\text{short}} | A=0) \neq 0$, unless $ \frac{Cov(X_1, X_2)}{Var(X_1)}  \left[ \mathop{\mathbb{E}}(X_1 | A=1) - \mathop{\mathbb{E}}(X_1 | A=0) \right] =  \left[ \mathop{\mathbb{E}}(X_2 | A=1)  - \mathop{\mathbb{E}}(X_2 | A=0) \right].$
\end{proposition}

\begin{proof}

\begin{align*}
    & \mathop{\mathbb{E}}(e_{\text{short}} | A=1) - \mathop{\mathbb{E}}(e_{\text{short}} | A=0)  \\
    &= \frac{Cov(X_1, X_2)}{Var(X_1)} \beta_2 \left[ \mathop{\mathbb{E}}(X_1 | A=1) - \mathop{\mathbb{E}}(X_1 | A=0) \right] -  \beta_2 \left[ \mathop{\mathbb{E}}(X_2 | A=1)  - \mathop{\mathbb{E}}(X_2 | A=0) \right] \\
    & \neq 0,
\end{align*}
and the inequality would hold unless a specific condition holds.
\end{proof}

We can see that the difference between the group-level mean prediction errors is a linear combination of the differences of expectations of feature $X_1$ between two groups, and the differences of expectations of feature $X_2$ between two groups.

\begin{proposition}
    $\mathop{\mathbb{E}}(e_{\text{short}} | A=1)  \Pr(A=1) + \mathop{\mathbb{E}}(e_{\text{short}} | A=0) \Pr(A=0) 
    = 0.$
\end{proposition}

\begin{proof}
\begin{equation}
\begin{aligned}
    \label{eq.linear.omitted.total.prob}
    0 = \mathop{\mathbb{E}}(e_{\text{short}})
    = \mathop{\mathbb{E}}(e_{\text{short}} | A=1)  \Pr(A=1) + \mathop{\mathbb{E}}(e_{\text{short}} | A=0) \Pr(A=0) 
\end{aligned}
\end{equation}
\end{proof}

In the following result, our definition of \emph{worst-case} is in the same sense as implied by the mini-max model in \cite{agarwal2019fair}. Specifically, if an optimal solution from their bounded group loss formulation simultaneously achieves maximum (absolute value) loss for every group, then we say this is the worst case for an optimal solution of a mini-max optimization problem.

\begin{corollary}
\label{cor:regression_omitted_worst_case}
 When two groups have the same size, this would lead to the worst-case bias of $ | \mathop{\mathbb{E}}(e_{\text{short}} | A=1)  - \mathop{\mathbb{E}}(e_{\text{short}} | A=0) | = 2|\mathop{\mathbb{E}}(e_{\text{short}} | A=1)|$.
\end{corollary}
\begin{proof}

Since two groups have equal size, $\Pr(A=1) = \Pr(A=0)$. 
From Eq.\ref{eq.linear.omitted.total.prob} we know that $\mathop{\mathbb{E}}(e_{\text{short}} | A=1)  + \mathop{\mathbb{E}}(e_{\text{short}} | A=0) = 0.$ 
Thus the statement is true.
\end{proof}




\section{Classification} \label{sec.classification}
In this section, we demonstrate how model mis-specification leads to bias in classification. We focus on cases where machine learning models generate risk scores, and take prediction error as the difference between the machine predicted score and true risk. We show that our main results in the previous section still hold in binary classification. Since most of the binary classification models do not have close-form solutions for estimation, we analytically prove the results in a restricted setup under probit model. Similar results can also be found in the Appendix for logit model. To complement the theory, we show that our main results hold under different data generating processes and machine learning models in Section \ref{sec.numerical}.

\subsection{Data generating process}
We denote the binary response variable as $Z \in \{0, 1\}$, and define $Y$ as the true probability of $Z=1$, \emph{i.e.}, $\Pr(Z=1|Y=y) = y$. $Y$ is a random variable given by Equation \ref{eq.true.data.generating.process}, and $h(\cdot)$ can take various functional forms. $Z$ is observed in the data, while $Y$ is unobserved. In many cases, machine learning algorithms produce risk scores, which are estimates of $Y$, based on the observed features and the binary responses. 

In probit model, $h$ is assumed to be a Cumulative Distribution Function (CDF):
\begin{align*}
Y = h(\mathbf{X})  = \Phi(\mathbf{X}^\intercal \bm{\beta}) = \Phi(\beta_0 + \beta_1 X_1 + \beta_2 X_2),
\end{align*}
where $\Phi$ is the CDF of the standard normal distribution. We assume that $X_1$ and $X_2$ are independent, and both are normally distributed: $X_1 \sim \mathcal{N}(\mu_1, \sigma_1^2), X_2 \sim \mathcal{N}(\mu_2, \sigma_2^2)$.
    
We further assume that $X_2$ has the same variance in the two groups, i.e., $Var(X_2| A=0) = Var(X_2| A=1) = Var(X_2) = \sigma_2^2$.
The probit model can be motivated as a latent variable model. Suppose there exists an auxiliary random variable
\begin{equation}
\begin{aligned}
    \label{eq.probit.true.dgp}
    Z^{*} = \mathbf{X}^\intercal \bm{\beta} + \epsilon,
\end{aligned}
\end{equation}
where $\epsilon$ follows a standard normal distribution. Then Z can be viewed as an indicator of the latent variable $Z^*$ being positive, i.e., $Z = \mathds{1}(Z^*>0)$.
To see that the latent variable model is equivalent to our risk model, we calculate probability of $Z=1$:
\begin{equation}
\begin{aligned}
    \label{eq.probit.link.formulation}
   \Pr(Z=1| \mathbf{X}) = \Pr(Z^* >0 | \mathbf{X})
     = \Phi(\mathbf{X}^\intercal \bm{\beta}) = Y.
\end{aligned}
\end{equation}

\subsection{Correctly specified model}
Similar to the case of linear regression, in probit model, if the model is correctly specified, then population-level mean prediction error is zero (Proposition \ref{prop:probit_pop_zero}), and algorithmic bias is zero (Corollary \ref{cor:probit_tau_zero}), even if the protected group is systematically different from the regular group in the observed features and labels.

\begin{proposition} \label{prop:probit_pop_zero}
     When the model is correctly specified, the mean prediction error in the population equals zero. That is if $Z^{*}= \gamma_0 + \gamma_1 X_1 + \gamma_2 X_2 + \xi$ where $\xi \sim \mathcal{N}(0,1)$ and is independent of $X_1$ and $X_2$, then mean prediction error  $\mathop{\mathbb{E}}(e^{\text{probit}}) = 0$, where $e^{\text{probit}}= \hat{Y} - Y $.
\end{proposition}

\begin{proof}


Given Eq. \ref{eq.probit.link.formulation}, we know $\gamma$ would satisfy the following:
\begin{align*}
    \Pr(Z=1| \mathbf{X}) =  \Phi(\gamma_0 + \gamma_1 X_1 + \gamma_2 X_2).
\end{align*}

Since we have the following:
\begin{align*}
    \Pr(Z=1 | \mathbf{X}) &=  \Pr(Z^* >0 | \mathbf{X}) = \Pr(\beta_0 + \beta_1 X_1 + \beta_2 X_2 + \epsilon >0 | \mathbf{X}) 
    = \Phi(\beta_0 + \beta_1 X_1 + \beta_2 X_2).
\end{align*}
Thus, the estimated coefficients would converge to the true coefficients, i.e., $\bm{\gamma} = \bm{\beta}$, and the prediction of $Y$ would be given as follows:
\begin{align*}
    \hat{Y} = \Phi(\gamma_0 + \gamma_1 X_1 + \gamma_2 X_2) = \Phi(\beta_0 + \beta_1 X_1 + \beta_2 X_2).
\end{align*}

Thus, the prediction error in this model would be as follows:
\begin{equation}
\begin{aligned}
\label{eq.probit.err.correct}
e^{\text{probit}} &= \hat{Y} - Y 
 = \Phi(\gamma_0 + \gamma_1 X_1  + \gamma_2 X_2) - \Phi(\beta_0 + \beta_1 X_1 + \beta_2 X_2) 
 = 0.
\end{aligned}
\end{equation}

Thus $\mathop{\mathbb{E}}(e^{\text{probit}}) = 0$.
\end{proof}

\begin{corollary} \label{cor:probit_tau_zero}
When the model is correctly specified, the difference between the expectations of the prediction error of the two groups equals zero. That is, $\tau = \mathop{\mathbb{E}}(e_{\text{short}} | A=1) - \mathop{\mathbb{E}}(e_{\text{short}} | A=0) = 0$.
\end{corollary}
This proof directly follows from Equation \ref{eq.probit.err.correct}, it is thus omitted.

\subsection{Model mis-specification and bias}
 Consider the case when the model is mis-specified with omitted variables ($X_2$ is omitted in the model). Here the estimation model is
 \begin{align*}
      Z^{*}  = \gamma_0 + \gamma_1 X_1 + \xi,
 \end{align*}
where $\mathbf{X}_{\text{short}} = (1, X_1)$, $\bm{\gamma}_{\text{short}}=(\gamma_0, \gamma_1)$ and $\xi$ follows a normal distribution. We denote the prediction error as $e_{\text{short}}$:
\begin{align*}
    e_{\text{short}} = \hat{Y}_{\text{short}} - Y.
\end{align*}

\begin{proposition}
     When the model is mis-specified with omitted variables, the mean prediction error in the population equals zeros, i.e., $\mathop{\mathbb{E}}(e_{\text{short}}) = 0$.
\end{proposition}
\begin{proof}

Recall that the true data generating process is given in Eq.\ref{eq.probit.true.dgp}, so we can rewrite that as follows:
\begin{align*}
    Z^{*}  &= \beta_0 + \beta_1 X_1 + u, \\
    u &= \beta_2 X_2 + \epsilon.
\end{align*}

Assume $X_2$ is independent of $X_1$ and follows a normal distribution $X_2 \sim \mathcal{N}(\mu_2, \sigma_2^2)$. Since $X_2$ and $\epsilon$ are independent, we now have:
\begin{align*}
    u = \beta_2 X_2 + \epsilon \sim \mathcal{N}(\beta_2 \mu_2, 1+\beta_2^2 \sigma_2^2).
\end{align*}

And we know that $u$ is independent of $X_1$. Thus we have:
\begin{align*}
    \Pr(Z=1 | X_1) &= \Pr( Z^{*} >0 |X_1) = \Pr(\beta_0 + \beta_1 X_1 + u >0|X
    _1) 
    = \Phi \Big( \frac{ \beta_0 + \beta_2 \mu_2}{\sqrt{1+\beta_2^2 \sigma_2^2}} + \frac{ \beta_1}{\sqrt{1+\beta_2^2 \sigma_2^2}} X_1  \Big).
\end{align*}

Thus we have:
\begin{align*}
    \gamma_0 = \frac{ \beta_0 + \beta_2 \mu_2}{\sqrt{1+\beta_2^2 \sigma_2^2}}, \quad
    \gamma_1 = \frac{ \beta_1}{\sqrt{1+\beta_2^2 \sigma_2^2}}.
\end{align*}

Thus, the prediction error in this model would be as follows:
\begin{equation}
\begin{aligned} \label{eq.probit_pred_error}
    e_{\text{short}} = \hat{Y}_{\text{short}} - Y = \Phi(\gamma_0 + \gamma_1 X_1 ) - \Phi (\beta_0 + \beta_1 X_1 + \beta_2 X_2 ).
\end{aligned}
\end{equation}

The mean prediction error in population is
\begin{align*}
    \mathop{\mathbb{E}}(e_{\text{short}}) &= \mathop{\mathbb{E}}(\Phi(\gamma_0 + \gamma_1 X_1 )) - \mathop{\mathbb{E}}(\Phi (\beta_0 + \beta_1 X_1 + \beta_2 X_2 )) \\
    &= \Phi(\frac{\gamma_0 + \gamma_1 \mu_1}{\sqrt{1+\gamma_1^2 \sigma_1^2}}) - \Phi(\frac{\beta_0 + \beta_1 \mu_1 + \beta_2 \mu_2}{\sqrt{1+\beta_1^2 \sigma_1^2 + \beta_2^2 \sigma_2^2}}) = 0
\end{align*}
\end{proof}

\begin{proposition}
When the model is mis-specified with omitted variables, the group-level mean prediction errors would not equal zero, unless for the omitted feature, its conditional expectation (on $A$) is the same for $A=0$ and $A=1$. That is $ \mathop{\mathbb{E}}(e_{\text{short}} | A=a) \neq 0 $ where $a=0$ or $1$, unless $\mathbb{E}(X_2 |A=a) = \mathbb{E} X_2$ $\forall a $.
\end{proposition}
\begin{proof}
From Equation \ref{eq.probit_pred_error}, we know that
\begin{align*}
    \mathop{\mathbb{E}}(e_{\text{short}} | A=a) &= \mathop{\mathbb{E}}(\Phi(\gamma_0 + \gamma_1 X_1 )| A=a) - \mathop{\mathbb{E}}(\Phi (\beta_0 + \beta_1 X_1 + \beta_2 X_2 )|A=a) \\
    &=  \Phi(\frac{\beta_0 + \beta_1 \mu_{1a} + \beta_2 \mu_2}{\sqrt{1+\beta_1^2 \sigma_{1a}^2 + \beta_2^2 \sigma_2^2}}) - \Phi(\frac{\beta_0 + \beta_1 \mu_{1a} + \beta_2 \mu_{2a}}{\sqrt{1+\beta_1^2 \sigma_{1a}^2 + \beta_2^2 \sigma_{2}^2}}),
\end{align*}
where $\mu_{1a} = \mathbb{E}(X_1| A=a)$, $\mu_{2a} = \mathbb{E}(X_2| A=a)$, and $\sigma_{1a}^2 = Var(X_1| A=a)$.
Since $\Phi$ is a strictly monotonic function,  $\mathop{\mathbb{E}}(e_{\text{short}} | A=a) = 0$ if an only if
\[\frac{\beta_0 + \beta_1 \mu_{1a} + \beta_2 \mu_2}{\sqrt{1+\beta_1^2 \sigma_{1a}^2 + \beta_2^2 \sigma_2^2}} = \frac{\beta_0 + \beta_1 \mu_{1a} + \beta_2 \mu_{2a}}{\sqrt{1+\beta_1^2 \sigma_{1a}^2 + \beta_2^2 \sigma_{2}^2}},
\]
which implies $\mu_2 = \mu_{2a}$, i.e., $\mathbb{E}(X_2 |A=a) = \mathbb{E} X_2$ $\forall a $.
\end{proof}

\begin{proposition}

\label{thm:probit_omitted_tau_not_zero} 
When the model is mis-specified with omitted variables, the difference between the group-level mean prediction errors would not equal zero and the model is biased under our fairness notion, unless  $\mathbb{E}(X_2 |A=a) = \mathbb{E} X_2$ $\forall a $.
\end{proposition}
\begin{proof}
We know that by definition
\begin{align*}
    (\mathbb{E}X_2 - \mathbb{E}(X_2| A=0))(\mathbb{E}X_2 - \mathbb{E}(X_2| A=1)) \leq 0.
\end{align*}
That is, $(\mu_2 - \mu_{20}) (\mu_2 - \mu_{21}) \leq 0$. This implies that
\begin{align*}
    \mathop{\mathbb{E}}(e_{\text{short}} | A=0) \cdot \mathop{\mathbb{E}}(e_{\text{short}} | A=1) \leq 0.
\end{align*}
The equality holds only when $\mathbb{E}(X_2 |A=a) = \mathbb{E} X_2$ $\forall a $. Thus, when $\mathbb{E}(X_2 |A=a) \neq \mathbb{E} X_2$ $\exists a$, we have
\begin{align*}
    \tau = \mathop{\mathbb{E}}(e_{\text{short}} | A=1) - \mathop{\mathbb{E}}(e_{\text{short}} | A=0) \neq 0.
\end{align*}
\end{proof}



\begin{table}[!htbp]
\renewcommand{\arraystretch}{1.2}
\begin{center}
\small
\caption {Simulation results} \label{table.simulation}
\begin{tabular}{llllllll}
\toprule
DGP & Model & Features & $b(\phi)$ & $b(\phi, A=0)$ & $b(\phi, A=1)$ & $\tau(\phi)$\\
\midrule
Linear & Linear &  $X_1, X_2$ & $1.52\times 10^{-14}$ & -$1.14\times 10^{-3}$ & $1.14\times 10^{-3}$ & $2.28\times 10^{-3}$  \\  
& & $X_1$ &  $6.96\times 10^{-15}$ & -$1.01$ & $1.01$ & $2.02$  \\  
Logistic & Logistic &  $X_1, X_2$ & $4.11\times 10^{-4}$  & $2.55\times 10^{-4}$ & -$1.08\times 10^{-3}$ & -$1.33\times 10^{-3}$  \\  
& & $X_1$ &  $4.11\times 10^{-4}$ & -$0.179$ & $0.178$ & $0.358$  \\  
Probit & Probit &  $X_1, X_2$ & $2.03\times 10^{-4}$ & -$7.85\times 10^{-4}$ & $1.19\times 10^{-3}$ & $1.98\times 10^{-3}$ \\  
& & $X_1$ &  $1.84\times 10^{-3}$ & -$0.226$ & $0.223$ & $0.449$  \\  
Linear & Random &  $X_1, X_2$ & $6.87\times 10^{-4}$ & $9.52\times 10^{-5}$ & $1.27\times 10^{-3}$ & $1.18\times 10^{-3}$  \\  
& Forest & $X_1$ &  $1.29\times 10^{-4}$ & -$0.491$ & $0.491$ & $0.982$  \\  
Polynomial & Linear &  $X_1, X_2$ & $1.26\times 10^{-15}$ & $0.376$ & -$0.376$ & -$0.752$  \\  
& & $X_1$ &  -$1.21\times 10^{-14}$ & -$1.01$ & $1.01$ & $2.02$  \\  
\bottomrule
\end{tabular}
\end{center}
\end{table}

\section{Numerical simulation}
\label{sec.numerical}

Let us introduce the general framework of simulation here. Recall that we need to specify two types of data generating processes: one for (linear) regression, and the other for classification. For each simulation, we specify the mapping function $h$ between feature $\mathbf{X}$ and $Y$. We also specify the group assignment $A_i$ for each observation $i$. To simulate each observation $i$, we generate its feature vector $\mathbf{X}_i$ and its observed outcome $Y_i$ independently according to the following procedures.

For regression type of DGP (linear and polynomial), the procedures are as follows:
\begin{enumerate}
    \item First, we simulate a two-dimensional feature vector
    \begin{align*}
        \mathbf{X}_i |A_i=a \sim  \mathcal{N}(\bm{\mu}^{A_i=a}, \bm{\Sigma}^{A_i=a}).
    \end{align*}
    
    \item Next, we create the observed outcome according to Eq.\ref{eq.true.data.generating.process}, which is:
    \begin{align*}
         Y_i = h(\mathbf{X}_i),
    \end{align*}
    where $h(\mathbf{X}_i) = \beta_0 + \beta_1 X_{i,1} + \beta_2 X_{i,2} + \epsilon_i$ in linear model and $h(\mathbf{X}) = \beta_0 + \beta_1 X_{i,1}  + \beta_2 X_{i,2}  + \beta_3 X_{i,1} ^2 + \beta_4 X_{i,2}^2 + \beta_5 X_{i,1}  X_{i,2}  + \epsilon_i$. In both cases,  $\epsilon^i \overset{i.i.d}{\sim} \mathcal{N}(0,1)$ and is independent of $\mathbf{X}_i$.
\end{enumerate}

For classification type of DGP (probit and logit), the procedures are as follows:
\begin{enumerate}
    \item First, we simulate a two-dimensional feature vector
    \begin{align*}
        \mathbf{X}_i |A_i=a \sim \mathcal{N} (\bm{\mu}^{A_i=a}, \bm{\Sigma}^{A_i=a}).
    \end{align*}
    
    
    \item Next, we create the observed outcome according to Eq.\ref{eq.true.data.generating.process}, which is:
    \begin{align*}
         Y_i = h(\mathbf{X}_i),
    \end{align*}
    where $h(\mathbf{X}_i) =  \Phi(\beta_0 + \beta_1 X_{i,1} + \beta_2 X_{i,2} )$ in probit model and $h(\mathbf{X}_i) =  S(\beta_0 + \beta_1 X_{i,1} + \beta_2 X_{i,2} )$ in logit model.
    
    \item Finally, we simulated the observed outcome according to:
    \begin{align*}
        Z_i \sim \text{Bern}(Y_i).
    \end{align*}
\end{enumerate}

We set two groups to have the same size, each with $10,000$ observations. In Table \ref{table.simulation}, we show results for $\bm{\beta}$ where $\beta_0=-2, \beta_1 = 1, \beta_2 = 1$ in all settings. For $\beta_3, \beta_4$ and $\beta_5$ which only exists in Polynomial DGP, we let $ \beta_3 = 1, \beta_4 = 1, \beta_5 = -1$. And for the feature vector $\mathbf{X}_i$,
\begin{align*}
    \mathbf{X}_i |A_i=0 \sim \mathcal{N} \left(\begin{pmatrix}
    1 \\
    1
    \end{pmatrix},\begin{pmatrix}
    1 & 0.5 \\
    0.5 & 1
    \end{pmatrix}\right), \quad
     \mathbf{X}_i |A_i=1 \sim \mathcal{N} \left(\begin{pmatrix}
    1 \\
    3
    \end{pmatrix},\begin{pmatrix}
    1 & -0.5 \\
    -0.5 & 1
    \end{pmatrix}\right).
\end{align*}
Analogous results can be obtained for other $\bm{\beta}, \bm{\mu}, \bm{\Sigma}$ values thus we do not show them here. 

In Table \ref{table.simulation}, we first observe that when a model is correctly specified, for both regression and classification DGP, the mean prediction errors and bias are zero. This implies that if there is redundant encoding in the data, \emph{i.e.}, when features used in prediction are correlated with the sensitive attributes, the model needs not produce biased outcome. Secondly, we observe that when a model is mis-specified with omitted variables, for both regression and classification DGPs, their population-level mean prediction error is still close to zero. However, the group-level mean prediction errors are large, and also different across groups. More specifically, they are in the opposite directions, leading to large bias. Moreover, mis-specification due to omitted variables and mis-specification due to the mismatch of function classes both lead to this outcome. To summarize, our simulation results confirm our finding that model mis-specification can lead to uneven outcomes for different groups.

\section{Discussion}

Our work illustrates how model mis-specification can lead to uneven outcomes for different groups. We show that redundant encoding does not necessarily lead to biased predictions in correctly specified models. When a model is mis-specified with omitted variables, even though their population-level mean prediction error is still close to zero, the group-level mean prediction errors can be large and in opposite directions. Moreover, we provide closed-form expressions that characterize bias in predictions as functions of first and second moments of the joint distribution of features. The analytical results are limited to specific settings, though similar results can be shown numerically in more general settings. 

Data bias can occur due to historical reasons or sociotechnical reasons during the data creation process. Algorithms can exaggerate or propagate bias, and human intervention comes from pre-processing (\emph{e.g.,} data procurement) and post-processing (\emph{e.g.,} threshold setting). 
Our ultimate goal is to use this line of inquiry to design better algorithmic processes that takes data and human biases into consideration more explicitly. This first step allows us to start decoupling the impact of different causes of unfairness, namely from data, algorithms, and human intervention.


\newpage
\bibliography{reference}
\bibliographystyle{icml2021}

\end{document}